# Learning semantic Image attributes using Image recognition and knowledge graph embeddings

Ashutosh Kumar Tiwari
Department of Information Science, BMSCE, BANGALORE, INDIA
Email: reachatashutosh@gmail.com

Sandeep Varma Nadimpalli
Assistant Professor, Department of Information Science, BMSCE, BANGALORE, INDIA
Email: sandeepvarma.ise@bmsce.ac.in

*Abstract*— **Extracting structured knowledge from texts has traditionally been used for knowledge base generation. However, other sources of information, such as images can be leveraged into this process to build more complete and richer knowledge bases. Structured semantic representation of the content of an image and knowledge graph embeddings can provide a unique representation of semantic relationships between image entities. Linking known entities in knowledge graphs and learning open-world images using language models has attracted lots of interest over the years. In this paper, we propose a shared learning approach to learn semantic attributes of images by combining a knowledge graph embedding model with the recognized attributes of images. The proposed model premises to help us understand the semantic relationship between the entities of an image and implicitly provide a link for the extracted entities through a knowledge graph embedding model. Under the limitation of using a custom user-defined knowledge base with limited data, the proposed model presents significant accuracy and provides a new alternative to the earlier approaches. The proposed approach is a step towards bridging the gap between frameworks which learn from large amounts of data and frameworks which use a limited set of predicates to infer new knowledge.**

*Index Terms*— **Knowledge graph embeddings, Image attributes, semantic Information, Image recognition, entity embeddings, Convolutional Neural Nets, COCO dataset, GLoVe, YOLO**

## I. INTRODUCTION

Images have proved to be one of the richest sources of data and are also one of the most challenging data types to examine. Obtaining accurate visual recognition by using simple general-purpose systems is still an unsolved paradigm. To make meaningful predictions in cases where inputs are significantly different from the ones used during training, particularly, cases where images can be from an entirely new class of objects, it is essential to incorporate and utilize additional knowledge. In such scenarios, a solution to add extra knowledge for better accuracy can be established using textual data in the form of knowledge graphs [1,2]. Knowledge graphs extract semantic information from words and are a rich source of structured information[3]. The problem addressed by the paper makes use of word entity embeddings of these knowledge representations to learn the semantic link between the entities of the image.

The present work is a discourse on knowledge graphs as an alternative source of information for the visual recognition task. To further understand the significance and contribution made by the proposed approach in the present work, it is important to distinguish between the three approaches of visual recognition :

1. Zero-Shot scenario: Prediction about images that are not linked to any of the classes seen during the training time is made by the model. The availability of extra information about the novel classes during training significantly aids in making these predictions.

2. Standard classification scenario: a model makes predictions for the images that were available during training.

3. Open/Closed world scenario: the faced images neither belong to the seen classes nor is any extra information about their classes is available during the training.

Examining Open/Closed world scenarios using extra structured information via knowledge graphs can be a natural choice for extracting semantic meaning from images. This is an area which is little studied and forms the basis of our present work. There have been earlier attempts to solve the problem of learning semantic relationships among image entities and using them for image captioning[12,13]. Extending such earlier works, the paper carefully investigates and makes contributions in the space of finding semantic relationships between the image attributes with the help of the knowledge representation models, namely the knowledge graph embedding model. Furthermore, it also incorporates the external representational information from the image to aid in learning and helps to predict links between the recognized image entities/objects. This key finding of the paper can be extended for use in reasoning systems to decide on an action.

The present work seeks to bridge the gap of acquiring new knowledge from a limited set of already established premises[4] and learning from a large amount of data[5]. The solution consists of four stages. The first stage involves an object detection model[6] which would help learn the attributes/objects of the input image.

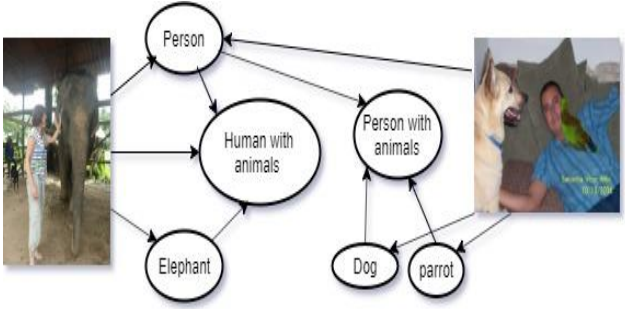

Fig.1. Custom user-defined knowledge base with image entities and their relationships

In the second stage, once these attributes of the images are available, the relationship between these entities is extracted from a ConceptNet [7] like, knowledge graph. However, for the current model, a custom knowledge base of entities and relationship has been built, where the Nodes represent the entities, and the Edges represents the relationships—an example of the custom user-defined knowledge base in giving in Fig.1. In the third stage, a knowledge graph embedding model is trained using this temporary database based on the work of *Liu. et al.'s* [8] Neural Association Model(NAM) with modifications. This serves as our baseline model. Further, the proposed model is then trained based on a Convolution Neural Network(CNN) to learn these relationships for successfully linking image entities. The aim of enhancing the learnings of the knowledge graph embedding model is fulfilled with the usage of explicit side- information from the image as input features, which would be in turn learned from a pre-trained VGGNet [9]. To summarize, the solution aims to understand how an image feature helps in enhancing the performance of the knowledge graph embedding model in terms of accuracy as compared to the baseline model where the image would be used as an input feature. Further, the paper discusses the datasets being used and the proposed solution architecture in detail in the sections ahead.

The rest of the paper is organized as follows. Section II includes the related literature for this work. Our proposed model architecture is detailed in Section III—a, where we explain the architectural design and setup of the research work. In Section III.B and Section III.C, we explain the detailed methodologies which are followed by our results of the baseline model and proposed model in Section IV. Section V includes the conclusion and discusses possible future work.

## II. RELATED WORK

The paper is positioned with respect to the most relevant research work in the field of image captioning, graph embeddings, knowledge bases, link prediction and semantic embedding[10,11]. There has been a wide range of work on image captioning problems in supervised settings which can be seen the work of Vinyals et al. [12] The work provides an end-to-end system to automatically describe the image contents and forms the basis of the idea of expressing semantic knowledge derived from image entities to a natural language like English. However, it does not address the use of unsupervised data from both texts and images to improve the image description. The present work carefully investigates these shortcomings to improve the image description task in hand. In work done by Frome et al. [13], named deep visual-semantic embedding(DeViSE) for the embedding of images in the vector space, a benchmark paradigm is established for providing a standard representation for unannotated text and images. Another similar work in this regard was done by Socher et al. [14] where the authors make such predictions for unknown classes through class labels. One of the most famous implementations of image captioning is by Vinyals and co. At Google called the Neural Image Caption Generator. This work demonstrated a very good solution for the problem of describing the content of an image by using a generative model based on Recurrent Neural Network's (RNNs). There is also a lot of research done on using semantic embedding for image captioning and image description.

One of the most interesting works that we came across was done in *Li et al*. [15] for the purpose of learning the semantic embeddings for visual recognition for the Multi-label classification task. This work models the intra- and inter-class relationships of image embeddings by exploiting discriminative and different constraints. However, it does little to address the scope of jointly learning the visual and textual embeddings. Another exciting work came from IBM research by *Lonij et al.* [16] where the authors have built a mutual embedding space for link prediction for an open-world visual recognition model using knowledge graphs. Our setup is very closely related to work done by *Lonij et al*. In the work done by IBM research, the authors develop a solution to a link prediction problem by using a knowledge graph model with embedded entities and encoded image representation as inputs. This work equally treats both image recognition(for known and unknown classes), as well as knowledge graph extension for predicting properties of image entities in open-world images. The evolution of the approach discussed in this work would require other data modes to be integrated into the system to automatically improve its systems' recognition capabilities. The current work envisions a model whose architectural design and setup are based on their implementation. In contrast, the previous work's implementation learns to solve a link prediction task using entity embeddings from the knowledge graph, in the current work the authors address and build their own small knowledge base with a closed-world assumption and use the image recognized entities and the image features as an external input to predict the actual class or relationship between the image objects.

III. METHODOLOGY

A. Model Architecture

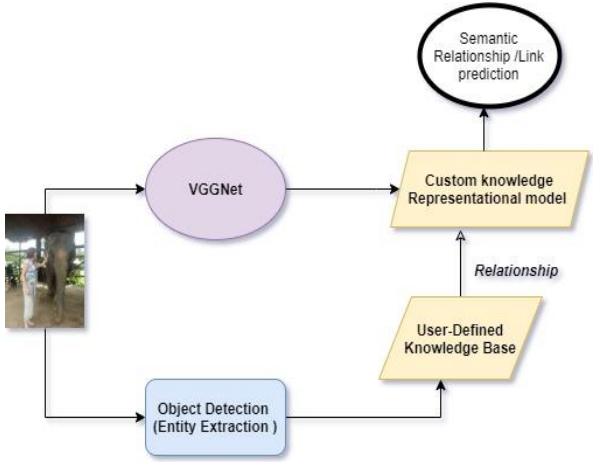

Fig.2. Proposed Model Architecture

Fig.2. Shows the proposed architecture for this research work. The architecture consists of blocks illustrating the various stages of establishing a relationship between the image entities using knowledge graph embeddings. The process starts with extracting entities in the input image through an object detection model such as the FastRCNN or YOLO [6]. The entities extracted from each image are then converted to vectors using a word2vec model, namely GloVe[17]. Here, each image object recognized is an entity such that the resulting vector is the average over the number of object entities recognized. This work makes use of a customized user-defined knowledge base, as represented in Fig.2. So that the relationship between the recognized object entities could be predicted from the knowledge base, a sample of which is shown in Fig.1. The architecture consists of a custom knowledge representation model that harbours the knowledge graph embeddings representing the knowledge bases. The specific input features are fed as a side-information, which would be in turn learned from a pre-trained VGGNet, to successfully predict the link or relationship between the image recognized entities and help in training. The paper establishes and compares the performance of a baseline model and the proposed model for the same task. The baseline model is based on Neural Association Model [8], where we replace the output layer by a softmax layer function for multiclass classification. The proposed model is based on a Convolution Network-based knowledge representation model such that along with the image we also use the prior information of the entity embeddings to predict the link present in the closed-world assumed knowledge base.

B. Knowledge Graph Embedding

For a unified representation of our custom-designed closed-world knowledge base and image inputs, a two-step approach is followed. First, the knowledge graph is used for the extracted entities to learn its representation. Subsequently, an image-vector mapping is learnt aided with the image representation extracted using the VGGNet.

These concepts are mathematically formalized in the following ways. Consider the finite set of knowledge represented as a knowledge graph $G$. A set of triplets of the for $\{(e_i, r, e_j) \in H, (e_i, e_j) \in E, r \in R\}$, where E is the set of entities in the vocabulary, and R is the set of type of relations between the extracted entities, model the knowledge graph. All such triplets are taken to be true if they occur in $G$ and are false if not in $G$. For a subset $H' \subset H$ of triples containing a subset of entities, $E' \subset E$, the task becomes finding the triples which are in $H$ but not in $H'$. The proposed model addresses the problem of having fewer data and predicts triples for unknown entities of $E$ that are not in $E'$. The labelled images are used for this purpose. An entity embedding $f: E \to V$ and an image embedding $g: I \to V$ where I is the set of all images and V is the vector space in $\mathbb{R}^d$.

The knowledge embedding model is trained using the set of triples $H'$. For a function mapping $f: E \to V$ which maps extracted entities to a vector space, we require a scoring function $S$ defined over the triplets which provide high scores to false triples and low scores to true triples. For this work, we used the neural tensor layer (NTL) architecture formulated in work done by *Socher et al.* as in (1),

$$S(f(e_h), r, f(e_t)) = \tanh\left(\left(f(e_h)^T w_r^{[1:k]}\right)^T f(e_t) + V_r[f(e_h)f(e_t)]^T + b_r\right) \quad (1)$$

Where $w_r^{[1:k]} \in \mathbb{R}^{d \times d \times k}$, $V_r \in \mathbb{R}^{k \times 2d}$ and $b_r \in \mathbb{R}^k$ are embedding parameters, k is the number of slices as in Socher et al. and $f(e_h)f(e_t)$ is the concatenation of the two vectors. Also, for a knowledge graph's each triple $(e_h, r, e_t)$, a relation $r$ which holds true links each head entity $e_h$ to tail entity $e_t$. In the proposed model, the image embedding $g: I \to V$ is modelled using a convolutional neural network—a set $D$ of labelled images $(l, e)$ where the labels $e \in E$ are entity types from the knowledge graphs. $g(l), l \in I$ is computed and mapped into the embedding space.

C. Experimental Setup

The current model combines the information obtained from the object recognition model along with semantic information from a word embedding model to determine the semantic attributes of the images. In order to achieve this goal, the authors decided to build their own dataset of 7584 images distributed across 12 custom defined image label class extracted from the COCO dataset[18]. The label categories for the images were defined based on variability and strength of the images in the particular class. The predefined set of image class labels along with their distributions in the training set is mentioned in Table.1. The images from the COCO image base have been selected with an effort on minimizing the effect of data skewness.

The object detection model used here is the state-of-the-art YOLO method[19]. YOLO uses a single convolutional neural network for both classification and localizing the object using bounding boxes[20]. The score for all the boxes detected is represented by a Tensor of shape 1x1. These scores are then sorted in descending order, and the non-max suppression algorithm is applied to remove any redundant boxes[21].

Table 1. This table shows the labels considered and their respective count of images in the training set

| Labels | Number of images |
|---|---|
| Human with animals | 1699 |
| Tennis racket | 997 |
| Baseball | 942 |
| Sportsball | 530 |
| Person snowboarding | 964 |
| Kitchen electronics | 31 |
| Living room | 86 |
| Traffic | 98 |
| Utencils | 209 |
| Person with bags | 623 |
| Animals | 926 |
| Human with Umbrella | 491 |

Furthermore, intersection over union (IOU)[22] is used to remove these redundant scores of bounding boxes. The remaining 3-5 boxes after this process which have the maximum score are considered for the prediction of classes of objects identified. For the model used in this research work, images to the YOLO-v3 model are not directly passed; rather, the images are first reshaped to size 416 X 416. Pre-trained weights were used for this YOLO model which was trained on the COCO dataset[23].

A VGGNet was used for the learning and extracting of the image representation using a vector size of 4096. The learned representation of these images was further used for predicting the link between the object extracted entities and the image itself. Thus, the authors used the image representation along with the object entity representation for the knowledge graph model for the link prediction task. Further, to create word embeddings, the authors used GloVe, coined from Global Vectors. GloVe is also a state-of-the-art model for distributed word representation[17]. This distributed word representation model is based on skip-gram based model for converting word to vector, where each spatial word position is considered with respect to the left and right context words associated with the current word. Thus, a vector embedding defines the syntactic as well as the semantic representation of the word. The embedding vector's dimension is currently set to 100; however, future experiments are open to exploration with a higher dimension of embedding vectors.

As a baseline model, the experimental setup involves a feedforward neural network with three hidden layers, and 2 Dropout layers to reduce the effects of over-fitting. The model's hidden units dimension is 128, 64, 64 and 12, where the last layer provides the classification into 12 different image class types or image entity type. The baseline neural network models the input entity based word embedding vector, where the entities are derived from the image to predict the image class.

Table 2. Baseline Model Architecture

| Layer Name | Hidden Units/Dropout rate |
|---|---|
| Dense-1 | 128 |
| Dropout | 0.5 |
| Dense-2 | 64 |
| Dropout | 0.5 |
| Dense-3 | 64 |
| Dropout | 0.5 |
| Dense-output | 12 |

For the proposed model, the experimental setup involves the use of a convolution neural network to process the compressed image representation, which is followed by a concatenation of compressed image representation and the word embedding. Further, the concatenated output is flattened and passed to a couple of dense layers. To compress the image representation given by VGGNet, we have used five convolution layers which are followed by max pool layers after every convolution layer. This output is concatenated and passed onto a single dense layer. This network further ends with the output dense layer along with the softmax activation function.

Table 3. Proposed Model Architecture

| Layer Name | Filter Size/Options |
|---|---|
| Conv-1 | 4x1 |
| Max Pool-1 | 2x1 |
| Conv-2 | 4x1 |
| Max Pool-2 | 2x1 |
| Conv-3 | 4x1 |
| Max Pool-3 | 2x1 |
| Conv-4 | 4x1 |
| Max Pool-4 | 2x1 |
| Conv-5 | 4x1 |
| Max Pool-5 | 2x1 |
| Concatenation | img_input+ emb_input |
| Flatten | - |
| Dense | 64 |

| Dense-output | 12 |

## IV. RESULT AND DISCUSSION

The following section recounts the results obtained for the baseline and the proposed model and substantiates essential properties of our method. Additionally, the limitations on the capabilities of the current research work have been discussed.

*A. Baseline Model*

Figure. Three and Figure. 4 shows the accuracy and loss value of our baseline model. This model only takes the word embedding as input. The model uses Adam's optimizer with a learning rate of 0.01. Categorical cross-entropy loss was used to predict the 12 class output. Architecture details of this model are given in Table 2. The model was run for 200 epochs with a validation split of 0.2, i.e. 20 per cent of total data was taken as validation data and rest was used for training. We have also performed ten-fold cross-validation, which receives an accuracy of 84.66%.

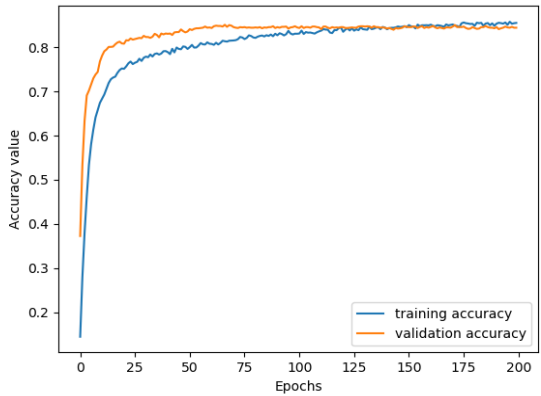

Fig.3. Accuracy for the baseline model

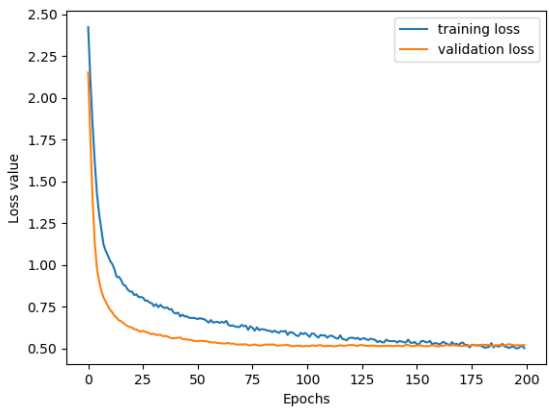

Fig.4. Loss for the baseline model

*B. Proposed model*

The details of the proposed model architecture are given in Table 3. The input is given in two parts first is the 4096 vector of images, and the second is 100 sized vector of word embedding. This model is also optimized using Adam optimizer, which has a learning rate of 0.01. We have also performed ten-fold cross-validation and have received an accuracy of 80.13%. The proposed model work with prior of the embedding vector given the encoded information from an image to predict the class of the images class label which can be stated in terms of knowledge graph model as the relation between the predicted entity object and the image.

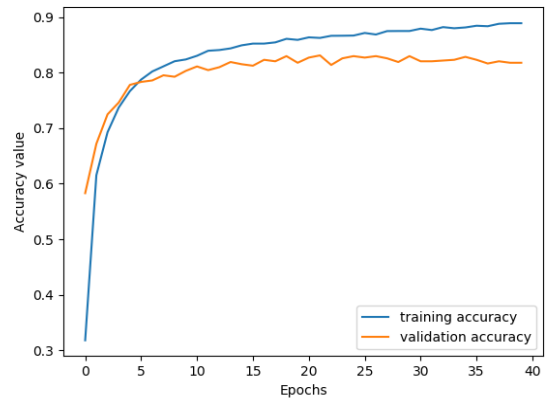

Fig.5. Accuracy for Proposed model

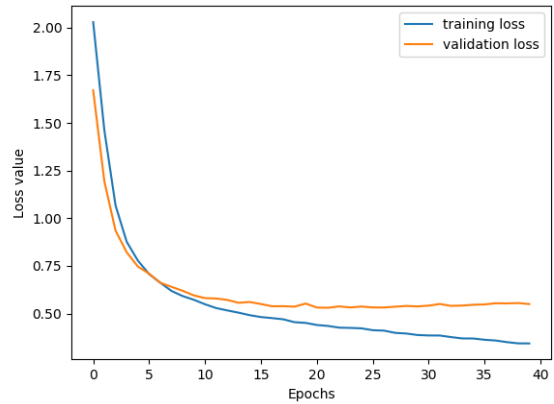

Fig.6. Loss for Proposed model

*C. Limitations of the Proposed model*

| Model | Accuracy(%) |
|---|---|
| Baseline | 84.66 |
| Proposed | 80.13 |

Table.4. Accuracy comparison of models

This section provides an overview of the validity and expectations that can be derived from the empirical results obtained and presented in this work. In the present setting, though the proposed approach displays low accuracy than the baseline approach, it's significant accuracy indicates that the proposed approach can be a new alternative for the problem in hand. Under changed settings of the ones discussed here, the accuracy is bound to increase. One of the significant limitations of the current model is with respect to the custom knowledge base that is defined for the proposed solution. The current knowledge base is based on closed-world assumptions. The knowledge base only defines 12 decided upon relationships between the image attributes and the corresponding images. Another limitation with respect to the amount of data that has been used to build the custom knowledge base for the creation of embeddings. The data is restricted to only 7584 images, due to which the proposed model does not give results that could be better than the proposed model. Also, since the authors have defined the own knowledge base, there could be gaps and discrepancies in defining the relationships and entities for this knowledge base which might affect the performance of the model leading to less accuracy.

## V. CONCLUSION

In this paper, we presented a convolutional neural network-based knowledge representation model for solving the link prediction task between images by establishing semantic relationships using knowledge graph embeddings. We compare our model with a baseline model which predicts the link using only the word embedding provided. The proposed model uses information from both, the image as well as the knowledge graph vector embedding to perform semantic learning of links of object entities recognized from an object recognition model. The proposed model performed the given task with an accuracy of 80.13% under the limitations of custom-designed closed-world knowledge base and a limited amount of training data. Furthermore, the output of the model makes an essential contribution and facilitates applications in decision-making systems where automated decision making based on images can be done.

The present research work has enough room for future investigation and experiment. In future, the authors would like to test this idea using a broader knowledge base with more classes and relationships. ConceptNet is one such example which can be considered for the same. Another future task can be to find a better way to select a subset of COCO dataset. This dataset should be in-line with the relationships found in the knowledge graph found from ConceptNet.

ACKNOWLEDGEMENT

The authors would like to extend their gratitude to all whose help and advice led to the completion of this work.

## Authors' Profiles

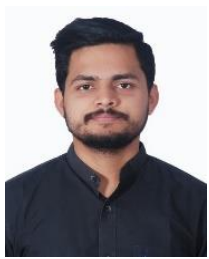

**Ashutosh Kumar Tiwari** is currently working as a Member Techincal Staff in Oracle Inc. He has completed his Bachelor of Engineering degree in Information Science and engineering from BMSCE, Banglore. He has interned at R&D labs of two major MNC's, Accenture and Epicor Incorporation and worked on Deep learning and reinforcement learning algorithms to drive business goals. He is an active member of Google developer group Bengaluru. His research interests include ML, DL, Data Engineering and Cloud Computing.

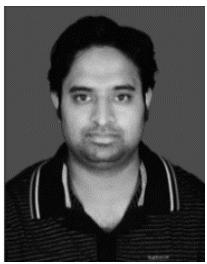

**Sandeep Varma Nadimpalli** is currently working as Assistant Professor since 2014 in the Department of Information Science and Engineering, B.M.S College of Engineering. He received his B.Tech. Degree in Information Technology from JNTU Hyderabad, Telangana, India in 2007. He received his M.Tech. From Andhra University in 2009 and his PhD in computer science and systems engineering from Andhra University in 2015. He also worked as a Junior Research Fellow (Professional) from 2010 to 2011 and later worked as Senior Research Fellow from 2011 to 2013 at Andhra University. His research interests include Data engineering, Data Privacy, Cloud Computing and Social Networks. He is a member of IEEE.